\documentclass[10pt,twocolumn,letterpaper]{article}

\usepackage[utf8]{inputenc}
\usepackage[T1]{fontenc}
\usepackage{lmodern}
\usepackage[margin=0.75in]{geometry}
\usepackage{xcolor,array}
\usepackage{graphicx}
\usepackage{amsmath}
\usepackage{amssymb}
\usepackage{mathtools}
\usepackage{float}
\usepackage{microtype}
\usepackage{booktabs}
\usepackage{multirow}
\usepackage{subfig}
\usepackage{cite}
\usepackage[colorlinks,urlcolor=blue,linkcolor=blue,citecolor=blue]{hyperref}
\usepackage[capitalize,noabbrev]{cleveref}

\begin{document}

%\title{TCN with Attention Model for Surrogate Semiconductor Fabrication Factory Simulation\\[0.6ex]}
\title{A Temporal Convolutional Network with Multi-Head Attention Model: A Surrogate for Parallel DEVS Semiconductor Fab Simulation}
% \large Archival Version of an Early Draft}

\author{%
Vamsi Krishna Pendyala$^{a}$, 
Hessam S. Sarjoughian$^{a,*}$, \\
Edward J. Yellig$^{b}$, 
Bharadhwaj Balasubramaniam$^{b}$
\\[1.2ex]
\small $^{a}$\textit{Arizona State University, Tempe, Arizona, USA}\\
\small $^{b}$\textit{Intel Corporation, Chandler, Arizona, USA}\\
\small $^{*}$Corresponding author. E-mail: hsarjou@asu.edu\\[0.6ex]%
}

% \date{Archived October 2026}

\twocolumn[
\begin{@twocolumnfalse}
\maketitle

% \begin{center}
% \fbox{\parbox{0.92\textwidth}{\small\textbf{Archival note.} This document is an archival copy of an early manuscript prepared during the initial development of the TCNA model. The experiments, figures, and tables reflect the early version of the MiniFab simulation data sets and have not been updated to later data or analyses. The text has been edited for grammar, consistency, and formatting only; it is not formatted for submission to any venue.}}
% \end{center}

\begin{abstract}
\noindent Semiconductor factories process many kinds of wafers organized into lots of various sizes and configurations. Different kinds of machines, each with its own operations, process wafers for hours to days, and the wafers undergo many incremental fabrication steps (with re-entry) and flows spanning days to weeks across connected machines. Fast and accurate predictions of throughput ($TH$) and turnaround time ($TA$) dynamics for individual machines and their combinations are essential for factory operations. To use machine learning methods for this purpose, this paper presents a Temporal Convolutional Network with Multi-head Attention (TCNA) architecture. It uses simulated datasets from a benchmark Parallel DEVS (PDEVS) semiconductor fabrication factory model. The factory processes configurable wafer lots with uniform and sinusoidal wafer arrival profiles and may undergo preventive maintenance. The wafer lot size, arrival profile, and factory maintenance policy are used as static covariates. TCNA is applied to a selected set of simulated discrete-event throughput and turnaround time trajectories. The $TH$ and $TA$ time series predicted by TCNA, evaluated with the common $MAPE$, $MSE$, and $R^2$ measures, show comparable or higher accuracy relative to other model architectures, with similar parameter counts and epoch times; the TCNA model size and execution time are similar to those of other architectures such as the Temporal Fusion Transformer. Although time series predictions are less accurate than high-resolution simulations, they can be deployed at a much lower computational cost. The TCNA achieves orders-of-magnitude speedup over PDEVS simulation with an acceptable loss of accuracy, enabling frequent product scheduling, capacity planning, and maintenance planning.
\end{abstract}

% \noindent\textbf{Impact statement---}Predicting the dynamics of systems with feedforward and feedback flows is needed for their analysis and design. Semiconductor manufacturing, characterized by stochastic dynamics driven by unpredictable events and timing, is an exemplar of such systems. Concepts, methods, and experiments for developing accurate models capable of fast simulation therefore stand to provide a basis for other kinds of event-driven systems. In this light, the TCNA architecture presented in this study uses simulated time series to capture short- and long-term temporal dependencies, contextual information, and static variables. TCNA provides more accurate predictions than several widely used time series forecasting models while requiring fewer parameters than the Temporal Fusion Transformer. This approach to predictive simulation enables faster deployment at a lower computational cost, making advanced predictive analytics suitable for near real-time decision support, and it can be applied to the predictive simulation of other event-driven systems, such as air, land, and ocean logistics and transportation.

% \medskip
% \noindent\textbf{Keywords---}Attention mechanism, machine learning, semiconductor process modeling, systems simulation, time series.

\vspace{1.5em}
\end{@twocolumnfalse}
]

\section{Introduction}
\label{section:introduction}

As stochastic, event-oriented systems, manufacturing factories involve many machines operating in mixed sequential and concurrent settings. In some factory segments, machines have only feedforward relationships, while in others they also have feedback relationships. Semiconductor factories have complicated dynamics and must operate under stringent time constraints on what machines do individually and collectively. Machines with different capabilities must accept a variety of wafer lots in different configurations and process them in different settings to produce different kinds of chips. 

A key characteristic of semiconductor fabrication is the long duration from the time raw wafers enter the factory until chips are shipped. Even though manufacturing chips typically takes several weeks to a few months, individual machines are expected to operate efficiently, given their workload capacity and resource availability (wafer lots and machines), at time scales of hours to a few days. Therefore, every machine in the process must meet time constraints determined in part by its upstream and downstream machines. Two essential metrics for manufacturing are throughput (the rate at which wafers are processed) and turnaround time (the total time required for raw wafers to become chips). 

Accurate and fast predictions of throughput and turnaround time for individual machines, and therefore for the whole factory, are of the utmost importance for efficient, cost-effective, and resilient factory operation. Over many years, models have been built using physics-based concepts, mathematical theories, methods, and tools. These models represent the structure and behavior of dynamical systems so that they can simulate the dynamics of components and their relationships with high accuracy and precision. Simulations of these models are integral to the development and operation of semiconductor manufacturing factories.

In recent years, researchers and practitioners have turned to machine learning (ML) as a means to understand factory dynamics by analyzing the large amounts of data routinely collected from individual machines. The high degree of automation used in smart manufacturing for individual machines and process flows allows high-resolution data collection at scale (e.g., \cite{GOLDMAN2021259,LINGITZ20181051,SUBRAMANIYAN2018533}). However, such data sets are closely guarded and are therefore not available to other researchers. Instead, suitable simulation models can be developed and simulated to obtain synthetic data sets, as is done in this research. A benchmark model of a semiconductor fabrication factory \cite{spier1995simulation} is used to develop discrete-event single-stage and multi-stage models. These models accept wafers of different types, configurations, and lot sizes, generated with uniform or sinusoidal arrival patterns. Machines can go offline (undergoing repair) for a finite period and then return to operation. The throughput and turnaround time trajectories are discrete-event and have a continuous time base. Wafers enter and exit machines at arbitrary time instants because the machines have stochastic dynamics. Such data sets can be converted to discrete-time trajectories for use in ML models \cite{WSC23}. 

This paper presents a Temporal Convolutional Network with Multi-head Attention (TCNA), a variant of the TCN architecture, for constructing time series models from simulation data sets obtained for a discrete-event semiconductor manufacturing factory model \cite{WSC23,ACIMS_neurIPS_2024}. A multi-head attention layer is used to obtain contextual information from historical values for predicting future values. The TCNA model predicts future values by taking into account the context of historical values along with the effects of static variables identified for the factory model, and it accurately reproduces the simulated throughput and turnaround time values. 

The simulations are developed and evaluated under low, medium, and high wafer volumes and different operational settings (e.g., machines that are temporarily unavailable). The prediction accuracy of TCNA is compared with that of the Autoformer \cite{wu2022autoformerdecompositiontransformersautocorrelation}, FEDformer \cite{zhou2022fedformerfrequencyenhanceddecomposed}, LSTM \cite{lstm}, N-BEATS \cite{oreshkin2020nbeatsneuralbasisexpansion}, TCN \cite{bai2018tcn}, and TFT \cite{lim2021temporal} models. Owing to the simplicity of the TCNA architecture, our efficiency analysis shows that TCNA needs fewer parameters and less memory than the other attention- and transformer-based models. TCNA predicts the $TH$ and $TA$ time trajectories with comparable or better mean absolute percentage error ($MAPE$), mean squared error ($MSE$), and $R^2$ score values.

\section{Semiconductor Manufacturing Factory Simulation Models}
\label{section:simulation_models}
The single-stage and multi-stage semiconductor factory models used in this research are developed using the Parallel Discrete Event System Specification (PDEVS) formalism \cite{chow1994parallel} and simulated using the DEVS-Suite simulator \cite{DEVS-Suite7}. An 8-stage cascade MiniFab factory model, composed of single-stage components, generates the simulation scenarios detailed in Section~\ref{section:experiments}. Each single-stage model has diffusion, implantation, and lithography segments, as defined for the benchmark Intel semiconductor factory called MiniFab \cite{spier1995simulation}. The diffusion stage receives raw wafers from three generators and partially processed wafers from the lithography stage. The implantation stage receives partially processed wafers from the diffusion and lithography stages. The lithography stage receives partially processed wafers from the implantation and diffusion stages. In the last step, the lithography stage completes the processing of the wafers received from the diffusion stage. The interval between two (or more) wafers arriving at (or leaving) a stage may range from minutes to multiple hours, which is a distinct characteristic of semiconductor factories.

The diffusion model has a dispatcher with a queue and two machines that independently perform the same loading, processing, unloading, and transportation of wafer batches. Similarly, the implantation and lithography models consist of components that receive, process, and forward wafers to other machines, each operating under specific timing constraints. Single-stage factories are connected in a cascade fashion to form multi-stage semiconductor fabrication factories. Wafers are organized into different batch configurations and processed through a six-step feedforward and feedback process encompassing the diffusion, implantation, and lithography stages. Discrete-event data from machines and dispatchers are collected by transducers without disrupting their operations. The time resolution for loading, processing, unloading, and transporting wafers is in minutes, with 10\% stochasticity. The 8-stage factory model is created by cascading eight single-stage factories, where the output of each upstream stage becomes the input of the downstream stage. The detailed design, simulation, data collection, and analysis of the PDEVS semiconductor fabrication factory models are described in \cite{WSC23}. These models are extended and simulated for multivariate time series analyses in \cite{ACIMS_neurIPS_2024,WSC24}.

\section{Related Work}
\label{section:related_work}
Discrete-event simulations (DES) are widely used to generate synthetic data when real-world data are limited or inaccessible \cite{chan2022generation}. For example, the single-stage MiniFab model has more than 70 simple and compound variable types (e.g., queues) \cite{WSC23}. Compared with data sets such as pedestrian counts and wind farms \cite{godahewa2021monash}, semiconductor factory data have unique characteristics. As discussed in \cite{ACIMS_neurIPS_2024}, simulation models play a vital role in time series analysis for manufacturing. Such synthetic data are strongly event-driven, requiring temporal ML models to consider contextual information for effective predictions \cite{WSC23}. In the domain of time series prediction, \cite{HUSSEIN2024100004} describes a hybrid model combining recurrent neural networks (RNNs) and genetic algorithms (GAs) to predict electricity consumption, and \cite{traffic2020} applies deep learning to traffic flow forecasting. Compared with the electricity and traffic domains, semiconductor manufacturing involves intricate wafer fabrication processes \cite{Lorenz1994, semiconductor_sim}, and its time series capture complex process flows influenced by multiple feedforward and feedback loops.

Transformer- and convolution-based architectures have gained prominence for time series forecasting \cite{zeng2023transformers, WSC24}. Temporal Fusion Transformers (TFT) \cite{lim2021temporal} effectively model static covariates, such as repair modes in semiconductor processes, and use multi-head attention to capture contextual dependencies. Although transformer models excel at long-term forecasting \cite{wen2022transformers}, their computational efficiency can be improved with convolutional operations \cite{ren2023transformer}. Lightweight models such as SparseTSF \cite{lin2024sparsetsfmodelinglongtermtime} and SAMformer \cite{ilbert2024samformerunlockingpotentialtransformers} demonstrate efficient adaptations of transformers, and ModernTCN \cite{luo2024moderntcn} and iTransformer \cite{liu2023itransformer} further optimize temporal feature extraction. For discrete-event time series, where both local transitions and global periodicity are critical, combining TCN and attention mechanisms can enhance temporal modeling. Unlike \cite{ZHEN2022111043}, which integrates TCN and attention in parallel, our architecture applies attention to the output of a TCN layer and feeds the result into a second TCN layer for richer feature extraction. In addition, gating mechanisms, inspired by \cite{choromanska2017gated}, dynamically control the flow of information and enhance predictive accuracy.

\section{Approach}
\label{section:approach}

As mentioned in Section~\ref{section:simulation_models}, the $TH$ and $TA$ series of a manufacturing factory are long (30,000 to 40,000 time steps), and forecasting them also depends on the static covariates associated with each series. Therefore, the problem can be formulated as follows:

\smallskip
\noindent\fbox{%
    \parbox{\dimexpr\columnwidth-2\fboxsep-2\fboxrule\relax}{%
        Given historical observations 
        $X = \{ x_{t-L}, \ldots, x_t \} \in \mathbb{R}^{L \times N}$, 
        where $N$ is the number of variates and $L$ is the look-back window size, 
        and a set of static covariates 
        $S = \{ s_1, \ldots, s_n \}, n \in \mathbb{N}$, 
        the aim is to predict $M$ future time steps 
        $Y = \{ y_{t+1}, \ldots, y_{t+M} \} \in \mathbb{R}^{M \times N}$.
    }%
}

\medskip

\noindent\textbf{Discrete-event-based forecasting:} Figures~\ref{fig:throughput_prediction_plots} and~\ref{fig:turnaround_prediction_plots} highlight the importance of contextual dependencies in these time series. The $TH$ and $TA$ time series exhibit \textit{non-stationary}, \textit{monotonically increasing} trends, with sharp variations influenced by historical values. In semiconductor fabrication, the intervals during which a machine is unavailable depend on factors such as usage, maintenance, and repair time. This local context can be expressed as $\text{Local Context} = f_l(X_{t-\Delta:t})$, where \( X_{t-\Delta:t} \) is the sequence of observations over the past \( \Delta \) time steps. In the same manufacturing system, the frequency of breakdowns may depend on seasonal trends, such as increased usage during peak production or reduced maintenance frequency over long intervals. This global context can be expressed as $\text{Global Context} = f_g(X)$, where \( X \) captures a long history of observations. Static factors such as machine type or operating policies significantly affect both contexts and are modeled as static covariates $S = \{ s_1, \ldots, s_n \}$, $n \in \mathbb{N}$. Events occur only at specific times (e.g., machine breakdowns and maintenance events); hence, observations are non-uniform over time, and their spacing is critical. Therefore, given historical observations \( X \), static covariates \( S \), and a target prediction horizon \( M \), the objective is to find a function \( g \) that minimizes the error \( \mathcal{L}(\hat{Y}, Y) \):
\[
\hat{Y} = g(f_g(X), f_l(X_{t-\Delta:t}), S).
\]
The loss function is the mean squared error:
   \[
   \mathcal{L}(\hat{Y}, Y) = \frac{1}{M} \sum_{i=1}^M \| \hat{y}_{t+i} - y_{t+i} \|^2,
   \]
where \( \|\cdot\|^2 \) denotes the squared error between the predictions \( \hat{y} \) and the true values \( y \). A simple architecture that captures sharp changes in the time series while retaining historical context serves this goal. The throughput and turnaround time obtained from the manufacturing simulation can be modeled as
\[
y_t = f(x_t, x_{t-1}, \dots, x_{t-L}; s_1, s_2, \dots, s_n) + \epsilon_t
\]
where \( \epsilon_t \) is a noise term. A rolling forecast is used to predict future values \(y_t\) iteratively, appending each prediction to the forecast sequence.

\subsection{TCN with Attention}
\label{subsection:temporal_convolutional_network_with_attention}
\begin{figure*}[!t]
    \centering
    \includegraphics[width=1\textwidth]{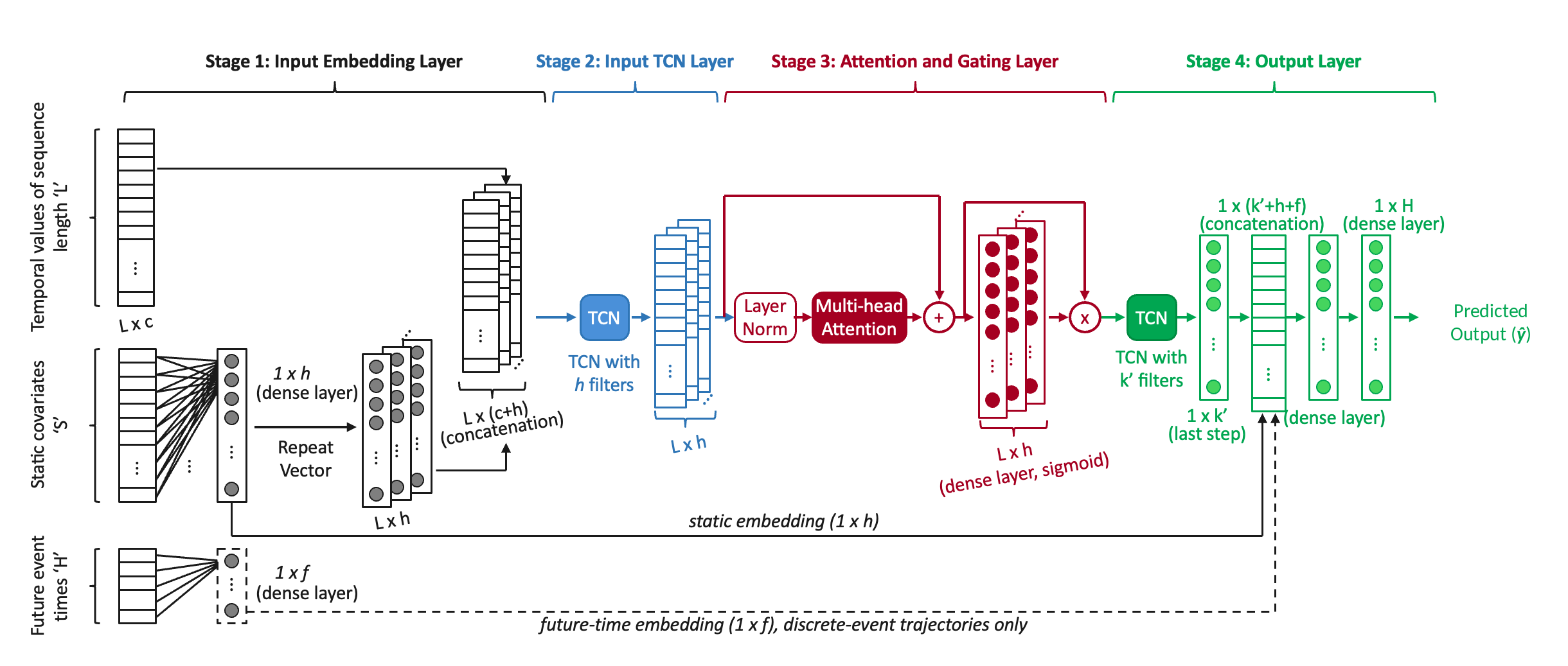}
    \caption{TCN with Multi-head Attention (TCNA) Model Architecture. The dashed future-time branch is an optional input for known-future (dynamic) covariates; it is not used in the experiments of this paper (see Section~\ref{section:conclusion}).}
    \label{fig:tcna}
\end{figure*}

TCNA combines dilated causal convolutions \cite{bai2018tcn}, which capture local temporal patterns such as step-to-step increments, with multi-head self-attention \cite{vaswani2017attention}, which relates all positions of the look-back window and thus captures the trend across it. The architecture is shown in Figure~\ref{fig:tcna} and is organized in four stages: the input embedding layer, the input TCN layer, the attention and gating layer, and the output layer. In contrast to residual formulations, which predict the change of the target from one step to the next, TCNA predicts the target values directly.

\textbf{Input Embedding Layer:} Let $\mathbf{X} \in \mathbb{R}^{L \times c}$ be the past window, which holds the $c$ dynamic input channels of the $L$ most recent observations; for the univariate $TH$ or $TA$ series considered here, these channels contain the scaled target values. Let $h$ be the static embedding width. The static covariates $\mathbf{s}$ (lot size, wafer types, wafer-arrival profile, and repair type) are embedded by a dense layer of width $h$, $\mathbf{s}_{emb} = \mathrm{ReLU}(\mathbf{W}_s \mathbf{s} + \mathbf{b}_s) \in \mathbb{R}^{h}$, repeated $L$ times, and concatenated with the past window:
\[
\mathbf{X}_{comb} = [\mathbf{X} \,\|\, \mathbf{1}_L \mathbf{s}_{emb}^{\top}] \in \mathbb{R}^{L \times (c+h)},
\]
where $[\cdot\|\cdot]$ denotes concatenation and $\mathbf{1}_L$ is a length-$L$ vector of ones. Every time step therefore carries the configuration of the factory that produced it.

\textbf{Input TCN Layer:} A TCN with $h$ filters, kernel size $\kappa$, and dilations $(1, 2, 4)$ or $(1, 2, 4, 8)$ is applied to $\mathbf{X}_{comb}$ with causal padding, residual and skip connections, and returned sequences, so that the output keeps the sequence length $L$. Let $u$ index positions in the input window, $j$ the output filters, $d$ a dilation factor, and $i$ the kernel offsets. With input vector $\mathbf{x}_{comb,u}$ and filter vector $\mathbf{w}_{i,j}$, each causal convolution computes
\[
h_j(u) = \mathrm{ReLU}\Big(\sum_{i=0}^{\kappa - 1} \mathbf{w}_{i,j}^{\top}\mathbf{x}_{comb,u-id} + b_j\Big), \quad j = 1, \ldots, h,
\]
where $h_j(u)$ is the output of filter $j$ at position $u$ and $b_j$ is its bias; causal padding supplies zeros for unavailable input positions, so each output uses only present and past inputs. Stacking dilations widens the receptive field of the layer. Writing $f_{TCN}$ for this transformation, the output is $\mathbf{X}_{TCN} = f_{TCN}(\mathbf{X}_{comb}) \in \mathbb{R}^{L \times h}$.

\textbf{Attention and Gating Layer:} The TCN output is layer-normalized \cite{ba2016layernorm} and passed through multi-head self-attention with $n_h$ heads. Let $a = 1, \ldots, n_h$ index the heads, $d_k$ be the key dimension, and $Q_a$, $K_a$, and $V_a$ be the query, key, and value projections of the normalized sequence. Then
\[
\text{Attention}_a = \text{softmax}\Big(\frac{Q_a K_a^{\top}}{\sqrt{d_k}}\Big) V_a, \quad a = 1, \ldots, n_h,
\]
where the softmax normalizes each row of attention scores to sum to one. The concatenated heads are projected back to width $h$ and added to the TCN output through a residual connection,
\[
\mathbf{Z} = \mathbf{X}_{TCN} + \mathrm{MHA}(\mathrm{LN}(\mathbf{X}_{TCN})),
\]
where $\mathrm{LN}$ denotes layer normalization and $\mathrm{MHA}$ multi-head attention. A sigmoid gate then scales each element of this representation:
\[
\mathbf{G} = \sigma(\mathbf{Z}\mathbf{W}_G + \mathbf{b}_G), \qquad \mathbf{X}_{gated} = \mathbf{G} \odot \mathbf{Z} \in \mathbb{R}^{L \times h},
\]
where $\sigma$ is the logistic sigmoid and $\odot$ the element-wise (Hadamard) product. Each entry of $\mathbf{Z}$ is thus multiplied by its own gate value in $(0,1)$, so the model can suppress positions and features that do not inform the forecast. This multiplicative gate is related to gated linear units \cite{dauphin2017gated} and to the gating of recurrent networks \cite{choromanska2017gated}, but here the representation is scaled by a sigmoid projection of itself.

\textbf{Output Layer:} A second TCN with $k'$ filters and dilations $(1, 2)$ or $(1, 2, 4)$ summarizes the gated sequence into its last step, $\mathbf{z}' \in \mathbb{R}^{k'}$. This vector is concatenated with the static embedding $\mathbf{s}_{emb}$ and mapped by two dense layers to the predicted target values:
\[
\hat{\mathbf{y}} = \mathbf{W}_o\, \mathrm{ReLU}\big(\mathbf{W}_u [\mathbf{z}' \,\|\, \mathbf{s}_{emb}] + \mathbf{b}_u\big) + \mathbf{b}_o \in \mathbb{R}^{H},
\]
where $H = M$ is the number of forecast steps ($H = 1$ for the one-step-ahead forecasts reported in Section~\ref{section:experiments}). Concatenating $\mathbf{s}_{emb}$ again at the output lets the factory configuration act directly on the prediction as well as through the encoded sequence. For known-future inputs, the dashed branch of Figure~\ref{fig:tcna} adds a future-time embedding $\mathbf{f}_{emb} = \mathrm{ReLU}(\mathbf{W}_f \mathbf{t}^{+} + \mathbf{b}_f) \in \mathbb{R}^{f}$ of the times $\mathbf{t}^{+}$ of the $H$ steps to be predicted, which is concatenated as $[\mathbf{z}' \,\|\, \mathbf{s}_{emb} \,\|\, \mathbf{f}_{emb}]$; this branch is not used in the present experiments.

\textbf{Local and global temporal structure:} The convolutional and attention stages capture the trajectory at different scales. A causal convolution combines each position with a fixed set of preceding positions through weights shared across the window, so it extracts local patterns of the target, such as step-to-step increments and the short bursts of completions of a batch, and stacking dilations widens this neighborhood. Self-attention instead weights every pair of positions in the window by the similarity of their representations, so each position can draw on any other position in a single layer, independently of their distance and of the receptive field. It can thereby relate the current state of the factory to the global course of the window, such as the start-up level or the rise towards a plateau. The residual connection in $\mathbf{Z}$ keeps the local convolutional features and adds this window-wide context, and the gate selects, per position and feature, how much of the combined representation reaches the output layer.

\subsection{Computational Efficiency}
\label{subsection:computational_efficiency}

To determine the efficiency of the model, we focus on the dominant terms in the computational cost of each component. For the input TCN layer with input $\mathbf{X}_{comb} \in \mathbb{R}^{L \times (c+h)}$, where $L$ is the look-back window and $c+h$ is the number of input features, the computation for $N$ stacked residual blocks with $h$ filters and kernel size $\kappa$ can be approximated as $\mathcal{O}(N \cdot L \cdot \kappa \cdot h \cdot (c+h))$. The time complexity of multi-head attention with $n_h$ heads of key dimension $d_k$ can be approximated as $\mathcal{O}(n_h \cdot (L^2 \cdot d_k + L \cdot h \cdot d_k))$. The total time complexity of the architecture can therefore be approximated as
\[
\mathcal{O}\big(N \cdot L \cdot \kappa \cdot h \cdot (c+h)\big)
+ \mathcal{O}\big(n_h \cdot (L^2 \cdot d_k + L \cdot h \cdot d_k)\big).
\]
Here, the local context is handled by the TCN layers, which leverage dilated convolutions for efficient local feature extraction, and the global context is captured by multi-head attention, which models relationships among all time steps. TCN layers are memory-efficient owing to convolutional sparsity, whereas attention layers use $\mathcal{O}(L^2)$ memory; hence, the look-back window length $L$ strongly affects the overall efficiency of the model.

\section{Experiments}
\label{section:experiments}

We conducted experiments with the TCNA model and other temporal ML models (see Table~\ref{table:model_performance}) to predict $TH$ (measured per minute) and $TA$ (measured in minutes) time series for the benchmark MiniFab model, with time $t$ defined in minutes. The $TH$ and $TA$ time series are approximations of their simulated discrete-event trajectory counterparts. Note that the scales of throughput and turnaround time differ by a factor of $10^6$. The $MAPE$, $MSE$, and $R^2$ values of TCNA and the other models are computed against the ground-truth PDEVS simulated data sets. The experiments also evaluate the efficiency of the TCNA model (in terms of the number of parameters and maximum memory) relative to the other temporal ML models. An ablation study shows the contribution of each TCNA component (see Tables~\ref{table:model_performance} and~\ref{table:ablation_study}).

\subsection{Experimental Setup}
\label{subsection:experimental_setup}
The temporal ML models (Autoformer, FEDformer, LSTM, N-BEATS, TCN, TCNA, and TFT) were trained and tested using the MiniFab simulation data sets. The MiniFab model processes wafer lots containing three wafer types, called Pa, Pb, and Tw, which arrive at 8-, 12-, and 24-hour intervals, respectively. Each wafer type is produced by its own generator, and the generated wafers follow either uniform or sinusoidal arrival patterns. The wafers are combined into five different kinds of batches before going through the first processing step in the diffusion and implantation stages. In contrast, the machines belonging to the eight lithography stages can process only one wafer at a time. Every machine can be configured to undergo one or more repairs during its use. A repair (during which the machine is under maintenance or offline) is triggered either according to the mean time between failures (MTBF) or after a certain number of lots have been processed. The wafer types, repair type, and lot size are used as static covariates for training the ML models.

\begin{table}[!htb]
\caption{Simulation Scenarios for Wafer Manufacturing}
\label{table:scenario}
\centering
\resizebox{\columnwidth}{!}{%
\begin{tabular}{cll}
\hline
\multicolumn{1}{l}{\textbf{Scenario}} & \textbf{Repair Type} & \textbf{Wafer Generation} \\ \hline
\textit{A} & No Repair & Sinusoidal \\
\textit{B} & No Repair & Uniform \\
\textit{C} & MTBF & Uniform \\
\textit{D} & Processing Steps & Uniform \\
\textit{E} & MTBF & Sinusoidal \\
\textit{F} & Processing Steps & Sinusoidal \\ \hline
\end{tabular}%
}
\end{table}

This configuration leads to the six scenarios in which the MiniFab model is simulated, as shown in Table~\ref{table:scenario}. The lot size is the total number of Pa, Pb, and Tw wafers used in the manufacturing process. Lot sizes are divided into three groups: small (60 to 108), medium (120 to 156), and large (168 to 204). Nine lot size and wafer type configurations were used for training in each scenario, totaling 54 time series. Similarly, three different batch configurations were used for testing in each scenario, totaling 18 time series. The models are thus trained on one set of configurations and tested on unseen wafer batch configurations. Although the time needed for the MiniFab to process all wafers differs among the small, medium, and large configurations, each of the 72 throughput and turnaround time series is set to 30,000 values spanning simulated $21.83$ days.

\begin{figure*}[htb!]
    \centering
    \includegraphics[width=0.65\linewidth]{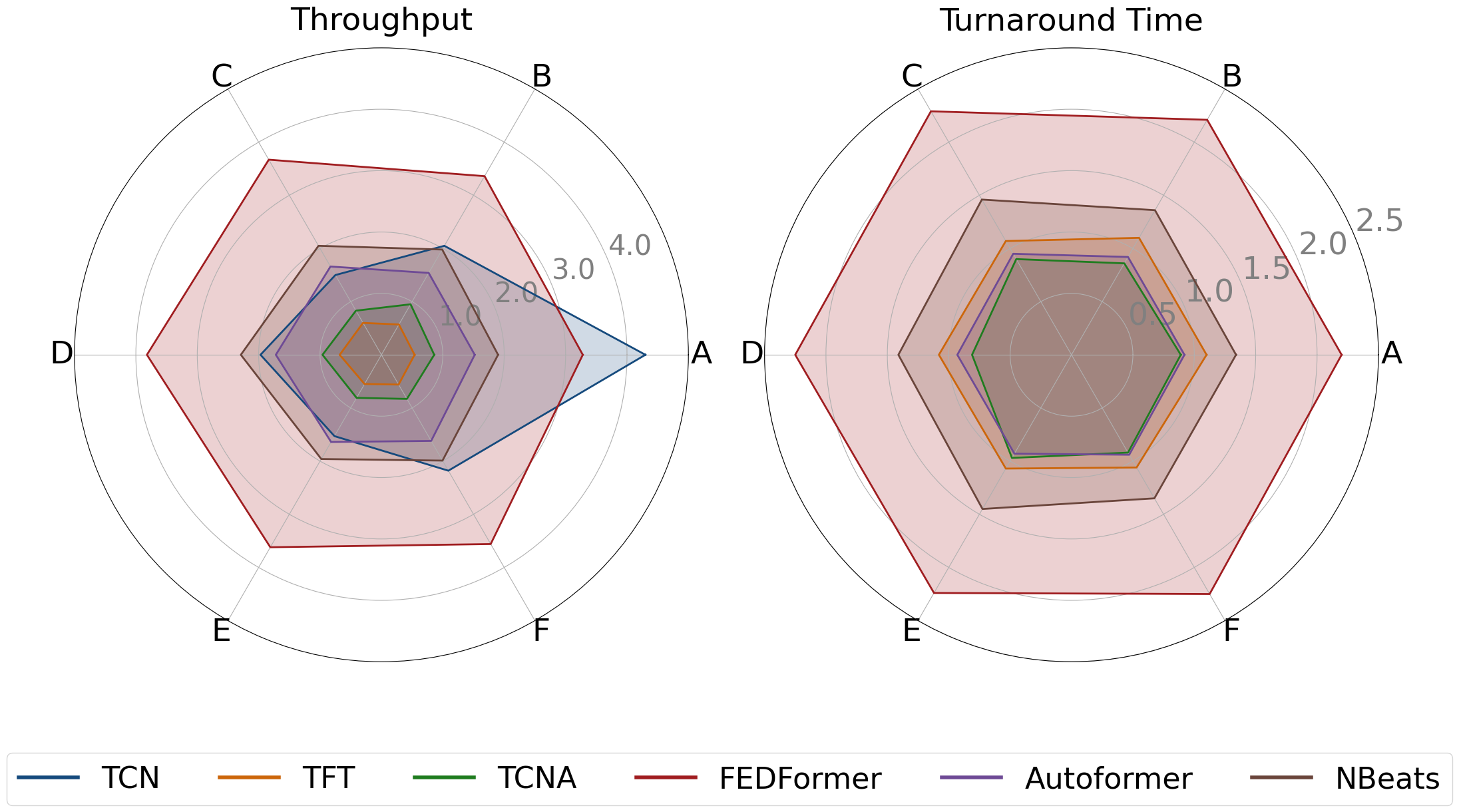}
    \caption{$MAPE$ Values for an 8-Stage Cascade Semiconductor MiniFab Factory}
    \label{fig:prediction_spider_plots}
\end{figure*}

\subsection{Prediction Measurements}
\label{subsection:prediction_measurements}

\begin{figure}[htb!]
    \centering
    \includegraphics[width=1\linewidth]{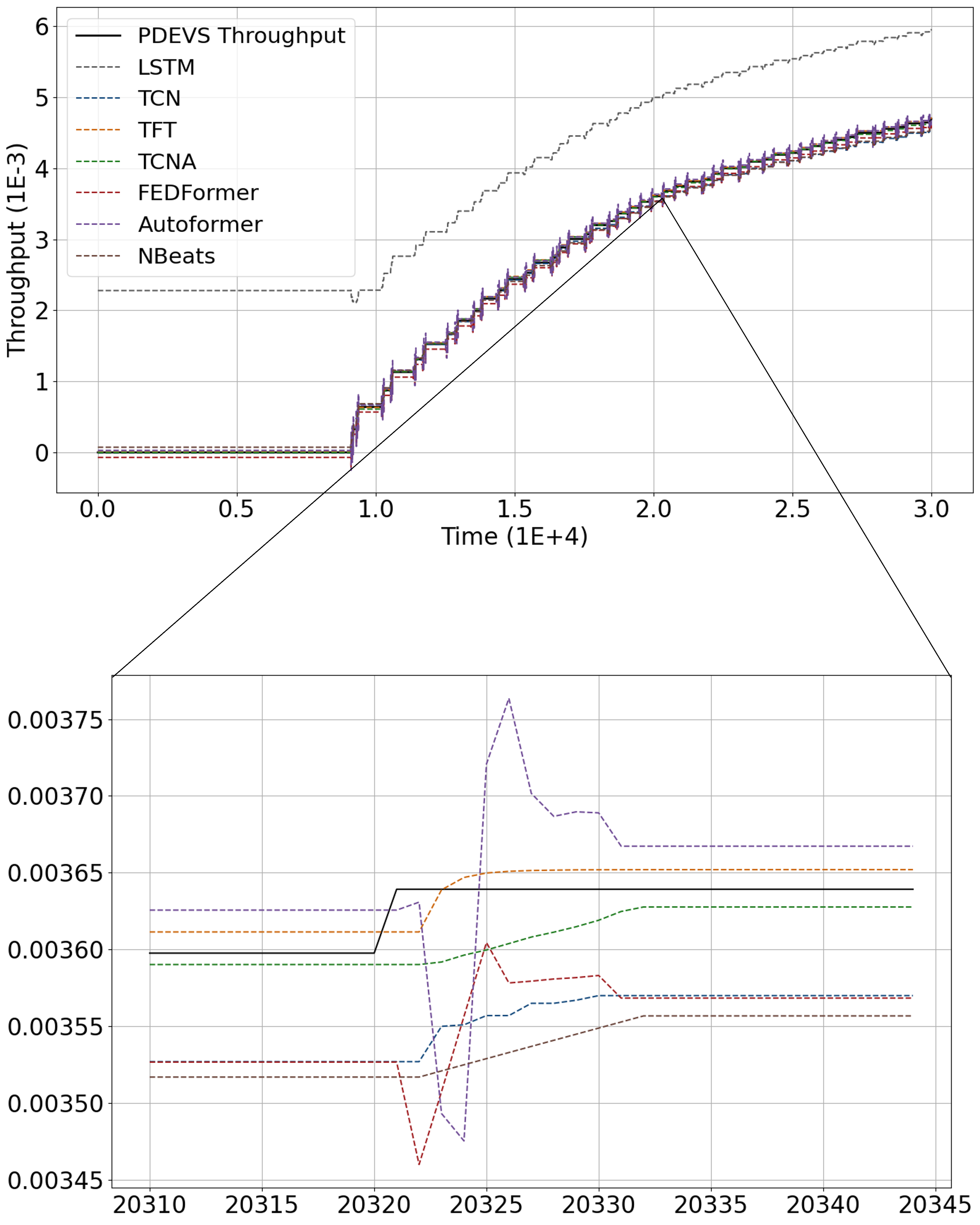}
    \caption{$TH$ Prediction Comparison for Medium Lot Size}
    \label{fig:throughput_prediction_plots}
\end{figure}

\begin{figure}[htb!]
    \centering
    \includegraphics[width=1\linewidth]{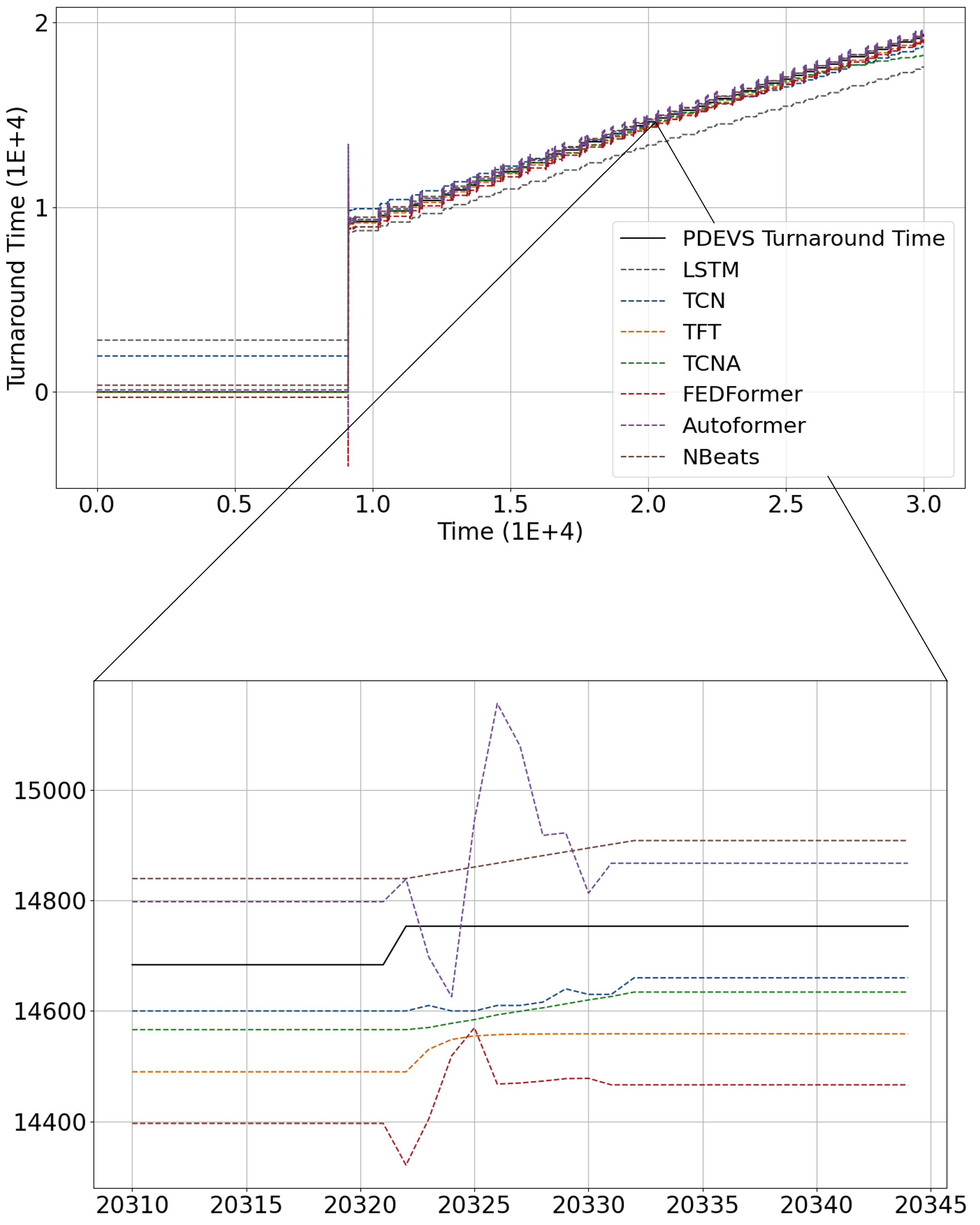}
    \caption{$TA$ Prediction Comparison for Medium Lot Size}
    \label{fig:turnaround_prediction_plots}
\end{figure}

The prediction accuracies of TCNA, TFT, TCN, Autoformer, FEDformer, N-BEATS, and LSTM are evaluated for throughput ($TH$) and turnaround time ($TA$) using input embeddings and static covariates. The $R^2$, $MSE$, and $MAPE$ metrics are summarized in Table~\ref{table:model_performance}, and the $MAPE$ values are visualized in Figure~\ref{fig:prediction_spider_plots}. For throughput, TFT achieves the lowest $MAPE$, followed by TCNA, whereas TCNA attains the best $MSE$ and $R^2$, with TFT ranking second; for turnaround time, this trend is reversed. For the medium lot size configuration (Pa = 72, Pb = 45, Tw = 27), Figures~\ref{fig:throughput_prediction_plots} and~\ref{fig:turnaround_prediction_plots} show that TCNA and TFT most closely track the MiniFab simulation trajectories. In contrast, LSTM and TCN do not capture the correct directional changes in throughput, while FEDformer, Autoformer, and N-BEATS follow the trends with reduced accuracy. All models use a 10-minute look-back window to predict one-step-ahead values, a design choice aimed at achieving the high short-horizon accuracy needed in semiconductor fabrication factories.

\begin{figure}[htb!]
    \centering
    \includegraphics[width=1\linewidth]{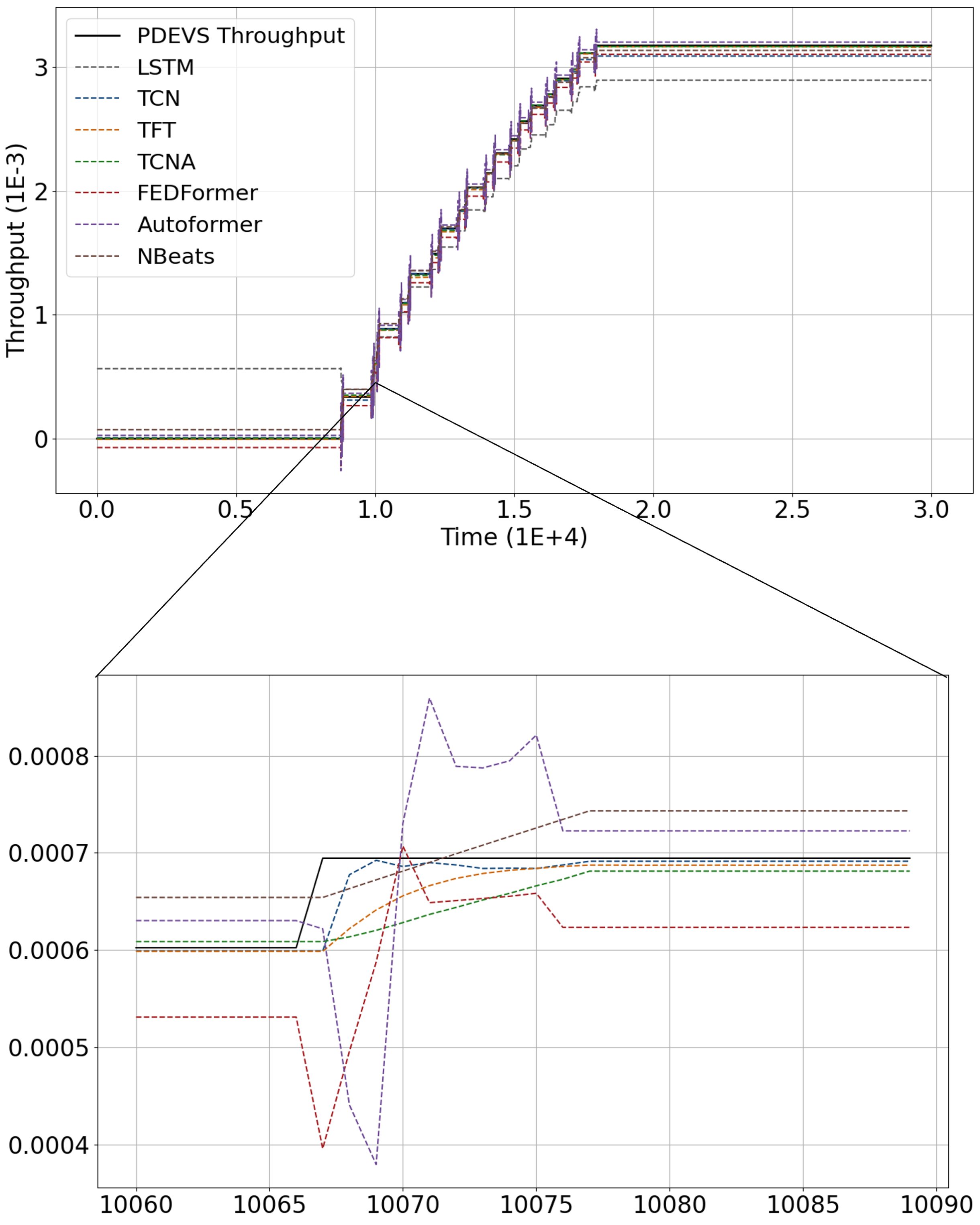}
    \caption{$TH$ Prediction Comparison for Small Lot Size}
    \label{fig:throughput_prediction_plots_small}
\end{figure}

\begin{figure}[htb!]
    \centering
    \includegraphics[width=1\linewidth]{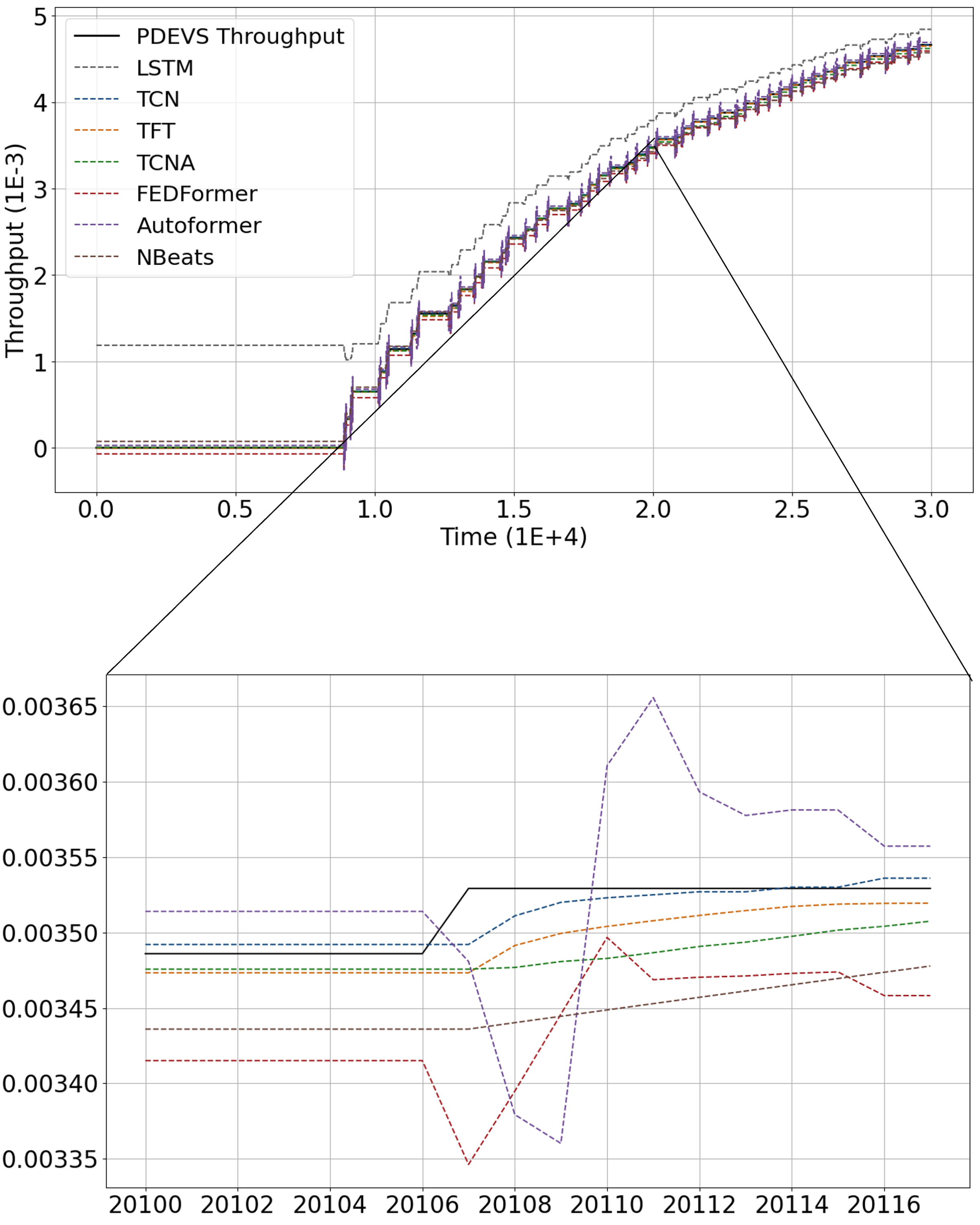}
    \caption{$TH$ Prediction Comparison for Large Lot Size}
    \label{fig:throughput_prediction_plots_large}
\end{figure}

\begin{figure}[htb!]
    \centering
    \includegraphics[width=1\linewidth]{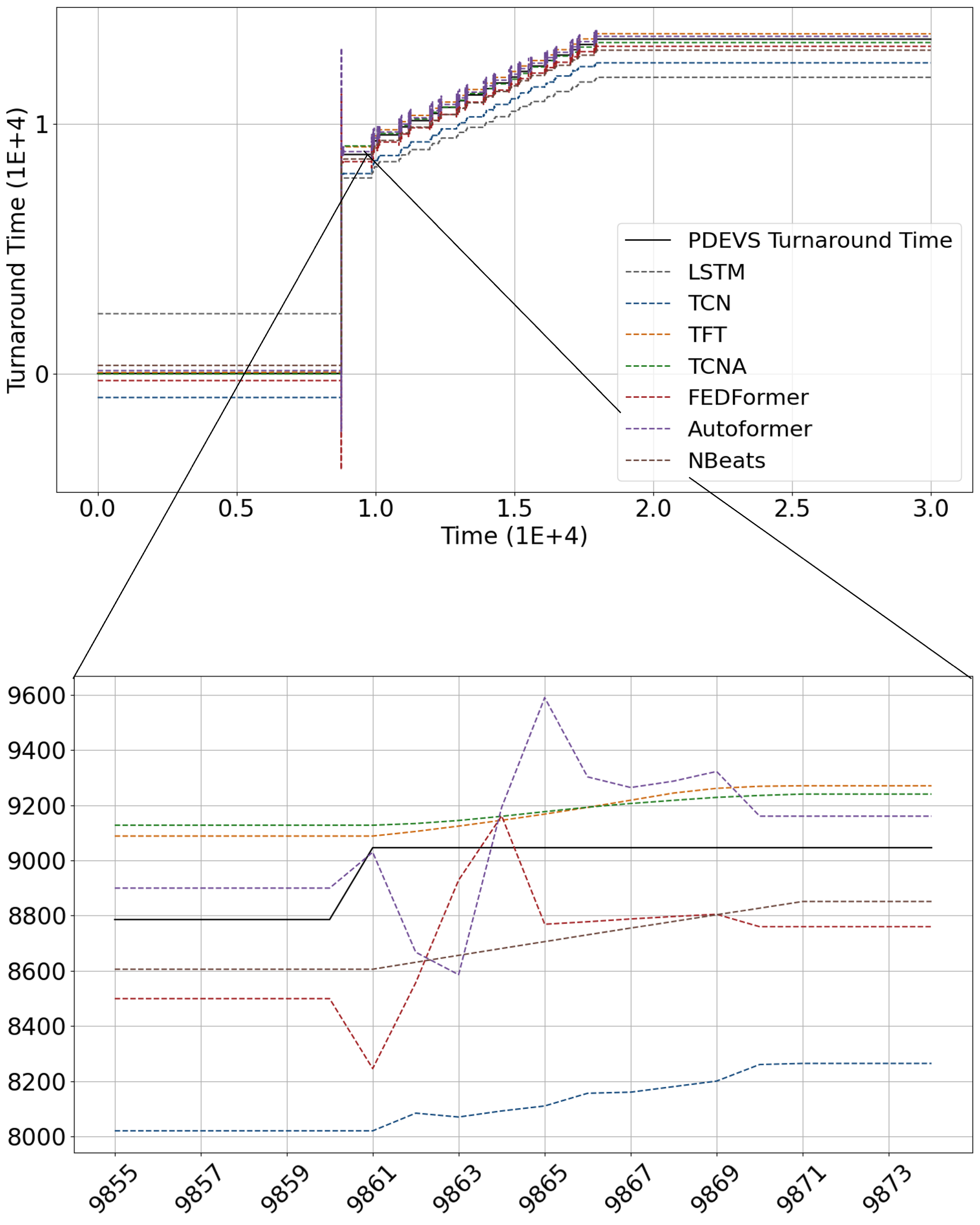}
    \caption{$TA$ Prediction Comparison for Small Lot Size}
    \label{fig:turnaround_prediction_plots_small}
\end{figure}

\begin{figure}[htb!]
    \centering
    \includegraphics[width=0.95\linewidth]{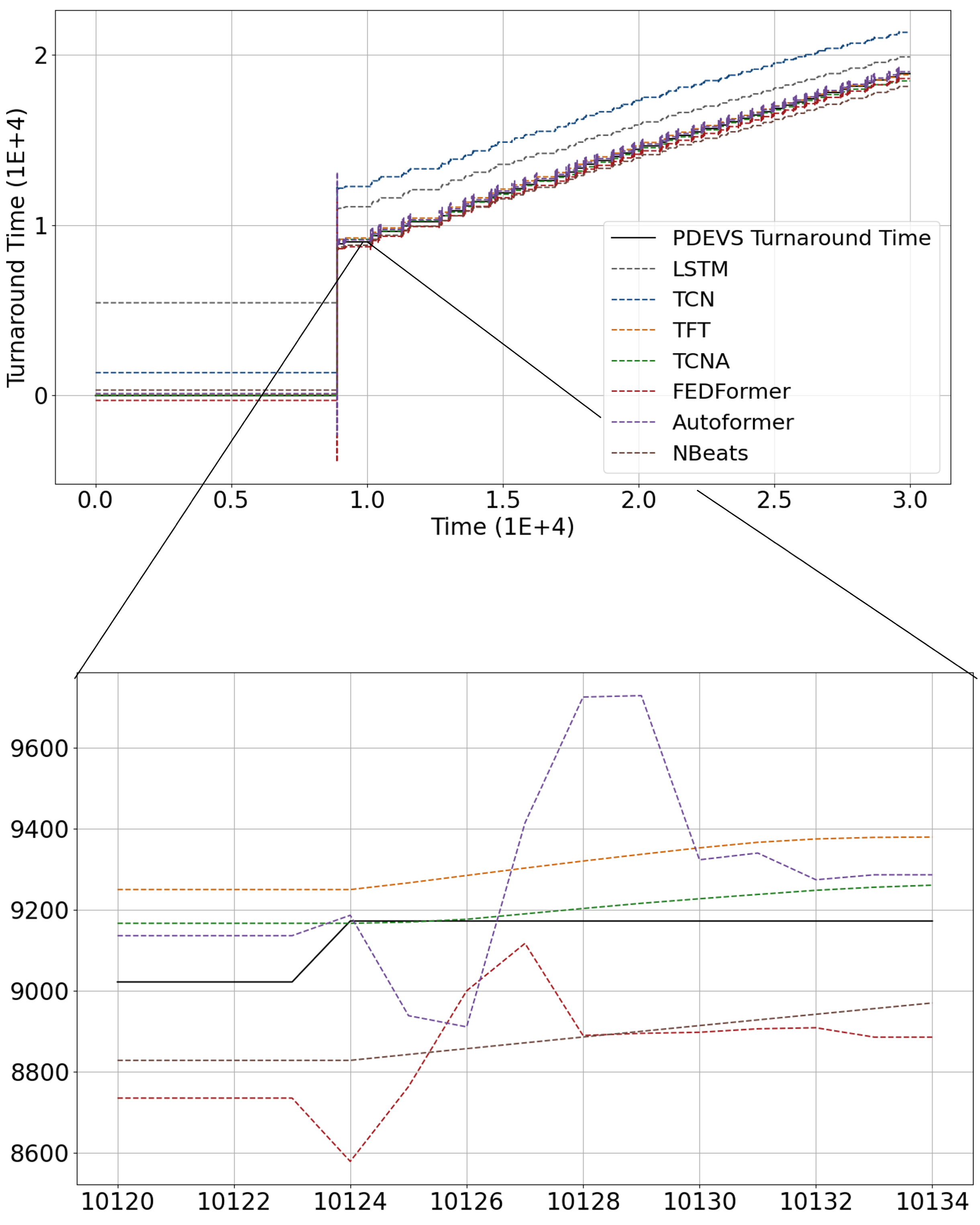}
    \caption{$TA$ Prediction Comparison for Large Lot Size}
    \label{fig:turnaround_prediction_plots_large}
\end{figure}

Further analysis highlights the impact of lot size on prediction behavior. For small lot sizes (Figures~\ref{fig:throughput_prediction_plots_small} and~\ref{fig:turnaround_prediction_plots_small}), shorter processing horizons and abrupt transitions increase sensitivity to local temporal dynamics, leading to delayed or oscillatory predictions from LSTM, TCN, Autoformer, and FEDformer, particularly near the first non-zero event (Figures~\ref{fig:throughput_first_non_zero} and~\ref{fig:turnaround_first_non_zero}). As lot size increases (Figures~\ref{fig:throughput_prediction_plots_large} and~\ref{fig:turnaround_prediction_plots_large}), the $TH$ and $TA$ trajectories become smoother and extend over longer time horizons, reducing short-term volatility but introducing cumulative drift, especially for turnaround time. Across all lot size regimes, TCNA maintains stable alignment with the PDEVS trajectories by capturing both local transitions and long-range dependencies, resulting in more consistent performance and reduced sensitivity to changes in batch configuration.

\begin{table*}[!t]
\caption{Model Performance Comparison: Average Values for Scenarios A--F}
\label{table:model_performance}
\centering
\resizebox{\textwidth}{!}{%
\begin{tabular}{ccllllllll}
\hline
\multicolumn{1}{l}{\textbf{Dataset}} & \multicolumn{1}{l}{\textbf{Scenario}} & \textbf{Metrics} & \textit{\textbf{LSTM}} & \textit{\textbf{TCN}} & \textit{\textbf{TFT}} & \textit{\textbf{TCNA}} & \textit{\textbf{FEDformer}} & \textit{\textbf{Autoformer}} & \textit{\textbf{N-BEATS}} \\ \hline
\multirow{18}{*}{\textit{\textbf{Throughput}}} & \multirow{3}{*}{\textit{A}} & \textit{$R^2$ Score} & 0.6754 & 0.9833 & 0.9998 & \textbf{0.9999} & 0.9984 & 0.9997 & 0.9990 \\
 &  & \textit{MSE} & 1.22E-07 & 4.00E-08 & 5.67E-10 & \textbf{4.07E-10} & 5.22E-09 & 1.14E-09 & 3.32E-09 \\
 &  & \textit{MAPE} & 32.59\% & 4.30\% & \textbf{0.54\%} & 0.86\% & 3.28\% & 1.52\% & 1.90\% \\ \cline{2-10} 
 & \multirow{3}{*}{\textit{B}} & \textit{$R^2$ Score} & 0.6810 & 0.9977 & \textbf{0.9998} & \textbf{0.9998} & 0.9983 & 0.9996 & 0.9989 \\
 &  & \textit{MSE} & 1.20E-06 & 6.07E-09 & \textbf{5.77E-10} & 6.43E-10 & 5.22E-09 & 1.15E-09 & 3.35E-09 \\
 &  & \textit{MAPE} & 32.56\% & 2.05\% & \textbf{0.57\%} & 0.95\% & 3.36\% & 1.54\% & 1.98\% \\ \cline{2-10} 
 & \multirow{3}{*}{\textit{C}} & \textit{$R^2$ Score} & 0.6060 & 0.9990 & 0.9997 & \textbf{0.9999} & 0.9980 & 0.9996 & 0.9989 \\
 &  & \textit{MSE} & 1.21E-06 & 2.45E-09 & 5.30E-10 & \textbf{2.20E-10} & 5.21E-09 & 1.14E-09 & 2.92E-09 \\
 &  & \textit{MAPE} & 34.88\% & 1.50\% & \textbf{0.60\%} & 0.83\% & 3.67\% & 1.66\% & 2.05\% \\ \cline{2-10} 
 & \multirow{3}{*}{\textit{D}} & \textit{$R^2$ Score} & 0.6320 & 0.9980 & 0.9998 & \textbf{0.9999} & 0.9981 & 0.9996 & 0.9989 \\
 &  & \textit{MSE} & 1.21E-06 & 4.24E-09 & 5.50E-10 & \textbf{2.33E-10} & 5.22E-09 & 1.13E-09 & 3.08E-09 \\
 &  & \textit{MAPE} & 38.69\% & 1.97\% & \textbf{0.68\%} & 0.96\% & 3.82\% & 1.72\% & 2.29\% \\ \cline{2-10} 
 & \multirow{3}{*}{\textit{E}} & \textit{$R^2$ Score} & 0.6010 & 0.9990 & 0.9998 & \textbf{0.9999} & 0.9980 & 0.9996 & 0.9989 \\
 &  & \textit{MSE} & 1.22E-06 & 2.48E-09 & 5.37E-10 & \textbf{1.97E-10} & 5.21E-09 & 1.13E-09 & 2.91E-09 \\
 &  & \textit{MAPE} & 33.89\% & 1.53\% & \textbf{0.55\%} & 0.81\% & 3.62\% & 1.64\% & 1.96\% \\ \cline{2-10} 
 & \multirow{3}{*}{\textit{F}} & \textit{$R^2$ Score} & 0.6230 & 0.9972 & 0.9998 & \textbf{0.9999} & 0.9981 & 0.9996 & 0.9989 \\
 &  & \textit{MSE} & 1.22E-06 & 6.22E-09 & 5.37E-10 & \textbf{2.13E-10} & 5.22E-09 & 1.14E-09 & 3.01E-09 \\
 &  & \textit{MAPE} & 37.92\% & 2.18\% & \textbf{0.56\%} & 0.83\% & 3.56\% & 1.62\% & 1.99\% \\ \hline
\multirow{18}{*}{\textit{\textbf{\begin{tabular}[c]{@{}c@{}}Turnaround \\ Time\end{tabular}}}} & \multirow{3}{*}{\textit{A}} & \textit{$R^2$ Score} & 0.9144 & 0.9730 & \textbf{0.9996} & 0.9991 & 0.9977 & 0.9993 & 0.9983 \\
 &  & \textit{MSE} & 3.67E+06 & 1.32E+06 & \textbf{2.11E+04} & 4.43E+04 & 9.98E+04 & 2.97E+04 & 7.26E+04 \\
 &  & \textit{MAPE} & 10.72\% & 6.51\% & 1.10\% & \textbf{0.89\%} & 2.20\% & 0.92\% & 1.34\% \\ \cline{2-10} 
 & \multirow{3}{*}{\textit{B}} & \textit{$R^2$ Score} & 0.9133 & 0.9696 & \textbf{0.9995} & 0.9992 & 0.9977 & 0.9993 & 0.9983 \\
 &  & \textit{MSE} & 3.67E+06 & 1.41E+06 & \textbf{1.99E+04} & 3.58E+04 & 9.93E+04 & 2.93E+04 & 7.20E+04 \\
 &  & \textit{MAPE} & 10.88\% & 7.09\% & 1.10\% & \textbf{0.86\%} & 2.21\% & 0.92\% & 1.36\% \\ \cline{2-10} 
 & \multirow{3}{*}{\textit{C}} & \textit{$R^2$ Score} & 0.9121 & 0.9656 & \textbf{0.9995} & 0.9993 & 0.9976 & 0.9993 & 0.9982 \\
 &  & \textit{MSE} & 3.47E+06 & 1.44E+06 & \textbf{1.88E+04} & 3.00E+04 & 9.74E+04 & 2.80E+04 & 7.07E+04 \\
 &  & \textit{MAPE} & 10.32\% & 7.43\% & 1.07\% & \textbf{0.90\%} & 2.29\% & 0.95\% & 1.46\% \\ \cline{2-10} 
 & \multirow{3}{*}{\textit{D}} & \textit{$R^2$ Score} & 0.9137 & 0.9639 & \textbf{0.9996} & 0.9993 & 0.9976 & 0.9993 & 0.9982 \\
 &  & \textit{MSE} & 3.50E+06 & 1.54E+06 & \textbf{1.91E+04} & 2.93E+04 & 9.84E+04 & 2.87E+04 & 7.16E+04 \\
 &  & \textit{MAPE} & 10.07\% & 7.66\% & 1.08\% & \textbf{0.81\%} & 2.25\% & 0.93\% & 1.41\% \\ \cline{2-10} 
 & \multirow{3}{*}{\textit{E}} & \textit{$R^2$ Score} & 0.9138 & 0.9674 & \textbf{0.9996} & 0.9991 & 0.9977 & 0.9993 & 0.9983 \\
 &  & \textit{MSE} & 3.57E+06 & 1.45E+06 & \textbf{1.93E+04} & 4.05E+04 & 9.81E+04 & 2.86E+04 & 7.08E+04 \\
 &  & \textit{MAPE} & 10.15\% & 7.22\% & 1.07\% & 0.97\% & 2.24\% & \textbf{0.93\%} & 1.45\% \\ \cline{2-10} 
 & \multirow{3}{*}{\textit{F}} & \textit{$R^2$ Score} & 0.9145 & 0.9661 & \textbf{0.9995} & 0.9991 & 0.9976 & 0.9993 & 0.9985 \\
 &  & \textit{MSE} & 3.50E+06 & 1.48E+06 & \textbf{1.92E+04} & 4.14E+04 & 9.78E+04 & 2.84E+04 & 6.90E+04 \\
 &  & \textit{MAPE} & 9.91\% & 7.35\% & 1.06\% & \textbf{0.92\%} & 2.25\% & 0.94\% & 1.35\% \\ \hline
\end{tabular}%
}
\end{table*}

Table~\ref{table:statistical_analysis} reports the paired $t$-statistics and Kolmogorov--Smirnov (KS) statistics for the throughput ($TH$) and turnaround time ($TA$) predictions in Scenario E for the small (Pa = 3, Pb = 42, Tw = 15; lot size = 60), medium (Pa = 72, Pb = 45, Tw = 27; lot size = 144), and large (Pa = 150, Pb = 24, Tw = 18; lot size = 192) lot sizes. Scenario E (repair type = MTBF, wafer generation = sinusoidal) has the most complex dynamics of all the scenarios (see Table~\ref{table:scenario}). The $t$-statistics reveal substantial variability across models and configurations, particularly for throughput, where large magnitudes indicate statistically significant differences between the model predictions and the PDEVS reference. Models such as LSTM and FEDformer consistently exhibit large positive or negative $t$-values across lot sizes, suggesting systematic bias in their predictions. In contrast, TCNA produces $t$-statistics among the closest to zero in most configurations (e.g., $10.1$ for medium-lot throughput, and $50.4$ and $68.8$ for small- and large-lot turnaround time), although not for large-lot throughput; this indicates a generally smaller mean deviation from the simulated trajectories. The KS statistics provide complementary insight into distributional similarity. For both throughput and turnaround time, TCNA, TFT, and Autoformer generally achieve lower KS values than LSTM and TCN, particularly in the medium and large lot size configurations. This indicates closer alignment between the predicted and simulated value distributions. Notably, the KS statistics are markedly lower for the medium and large lot sizes than for the small lot size, reflecting smoother system dynamics and improved distributional agreement for most models. Overall, the combined $t$-test and KS-test results indicate that, while several models capture global trends, TCNA achieves consistently strong agreement with the PDEVS reference in both mean behavior and distributional characteristics across lot size regimes.

\begin{table*}[!t]
\centering
\caption{Statistical Analysis for Scenario E}
\label{table:statistical_analysis}
\resizebox{\textwidth}{!}{%
\begin{tabular}{ccllllllll}
\hline
\multicolumn{1}{l}{} &
  \multicolumn{1}{l}{} &
   &
  \multicolumn{7}{c}{\textbf{Model Scores}} \\ \hline
\textbf{Metrics} &
  \multicolumn{1}{l}{} &
  \textbf{Lot Size} &
  \textit{LSTM} &
  \textit{TCN} &
  \textit{TFT} &
  \textit{TCNA} &
  \textit{FEDformer} &
  \textit{Autoformer} &
  \textit{N-BEATS} \\ \hline
\multirow{6}{*}{\textbf{$t$-statistic}} &
  \multirow{3}{*}{\textit{\textbf{Throughput}}} &
  \textit{Small} &
  3.37E+02 &
  2.14E+02 &
  1.84E+02 &
  9.79E+01 &
  1.02E+03 &
  -2.08E+02 &
  \textbf{8.96E+01} \\
 &
   &
  \textit{Medium} &
  -1.68E+03 &
  1.59E+02 &
  -2.12E+02 &
  \textbf{1.01E+01} &
  8.02E+02 &
  -1.51E+02 &
  1.73E+02 \\
 &
   &
  \textit{Large} &
  -4.07E+02 &
  \textbf{-1.20E+02} &
  1.43E+02 &
  1.76E+02 &
  8.42E+02 &
  -1.57E+02 &
  1.22E+02 \\ \cline{2-10} 
 &
  \multirow{3}{*}{\textit{\textbf{Turnaround Time}}} &
  \textit{Small} &
  1.12E+03 &
  1.75E+03 &
  -4.95E+02 &
  \textbf{5.04E+01} &
  4.72E+02 &
  -1.42E+02 &
  4.15E+02 \\
 &
   &
  \textit{Medium} &
  4.79E+02 &
  \textbf{-1.17E+00} &
  2.96E+02 &
  7.70E+01 &
  4.42E+02 &
  -1.32E+02 &
  -1.79E+02 \\
 &
   &
  \textit{Large} &
  -6.78E+02 &
  -1.79E+03 &
  -2.52E+02 &
  \textbf{6.88E+01} &
  4.58E+02 &
  -1.36E+02 &
  3.45E+02 \\ \hline
\multirow{6}{*}{\textbf{KS-statistic}} &
  \multirow{3}{*}{\textit{\textbf{Throughput}}} &
  \textit{Small} &
  6.33E-01 &
  5.94E-01 &
  \textbf{5.67E-01} &
  \textbf{5.67E-01} &
  5.94E-01 &
  5.68E-01 &
  \textbf{5.67E-01} \\
 &
   &
  \textit{Medium} &
  5.54E-01 &
  1.09E-01 &
  4.02E-02 &
  4.05E-02 &
  6.63E-02 &
  \textbf{3.98E-02} &
  8.60E-02 \\
 &
   &
  \textit{Large} &
  1.27E-01 &
  \textbf{5.02E-02} &
  5.25E-02 &
  5.25E-02 &
  6.45E-02 &
  5.20E-02 &
  7.89E-02 \\ \cline{2-10} 
 &
  \multirow{3}{*}{\textit{\textbf{Turnaround Time}}} &
  \textit{Small} &
  7.12E-01 &
  6.53E-01 &
  5.96E-01 &
  \textbf{5.69E-01} &
  5.98E-01 &
  \textbf{5.69E-01} &
  6.07E-01 \\
 &
   &
  \textit{Medium} &
  1.87E-01 &
  1.07E-01 &
  \textbf{4.36E-02} &
  1.12E-01 &
  5.43E-02 &
  5.01E-02 &
  5.21E-02 \\
 &
   &
  \textit{Large} &
  2.22E-01 &
  3.23E-01 &
  5.83E-02 &
  6.40E-02 &
  6.21E-02 &
  \textbf{5.46E-02} &
  1.05E-01 \\ \hline
\end{tabular}%
}
\end{table*}

\subsection{Early- and Late-Stage Trajectory Predictions of Throughput and Turnaround Time}
\label{subsection:early_late_dynamics}

\begin{figure}[htb!]
    \centering
    \includegraphics[width=0.95\linewidth]{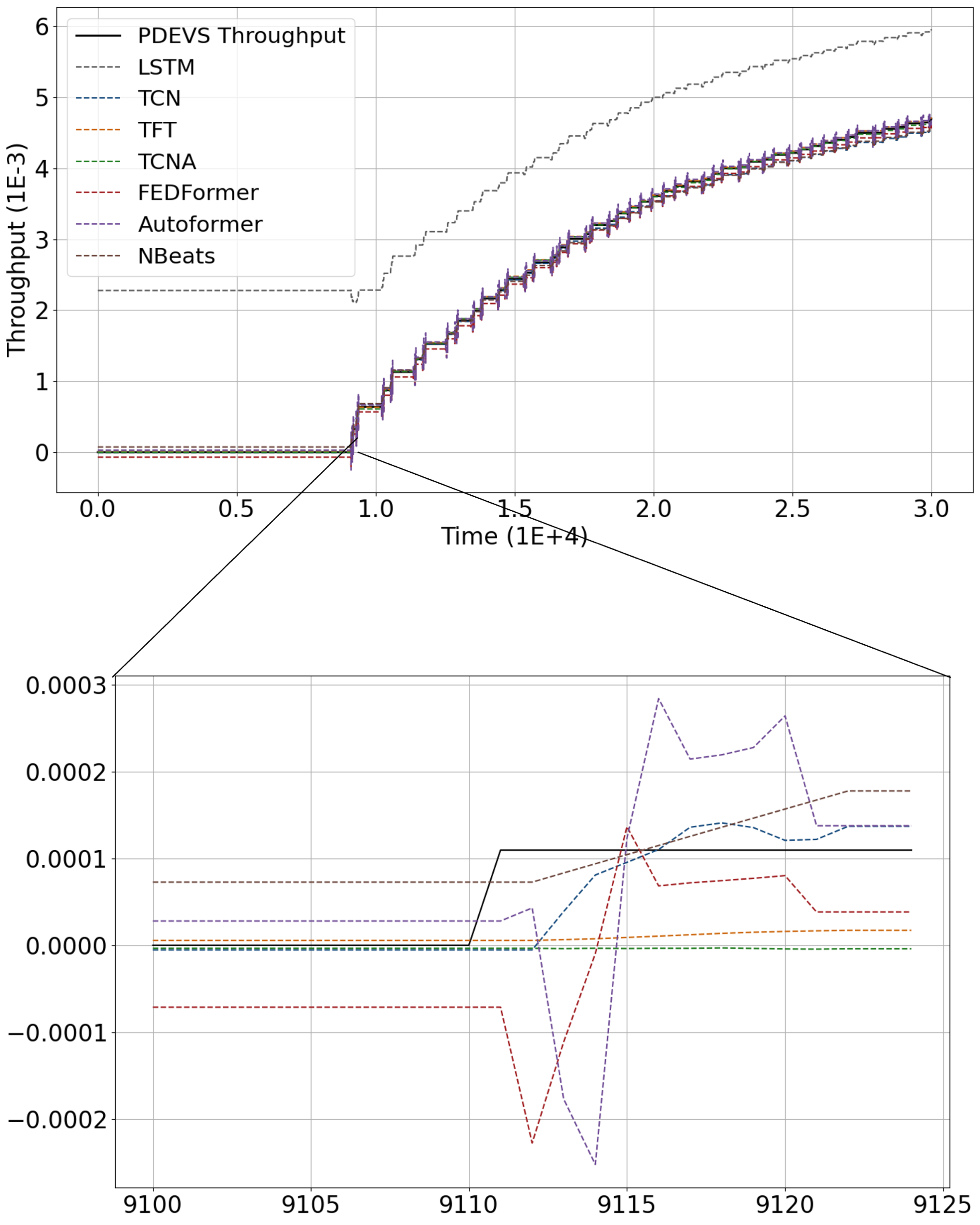}
    \caption{$TH$ Prediction Around the First Non-Zero Value for Medium Lot Size}
    \label{fig:throughput_first_non_zero}
\end{figure}

\begin{figure}[htb!]
    \centering
    \includegraphics[width=1\linewidth]{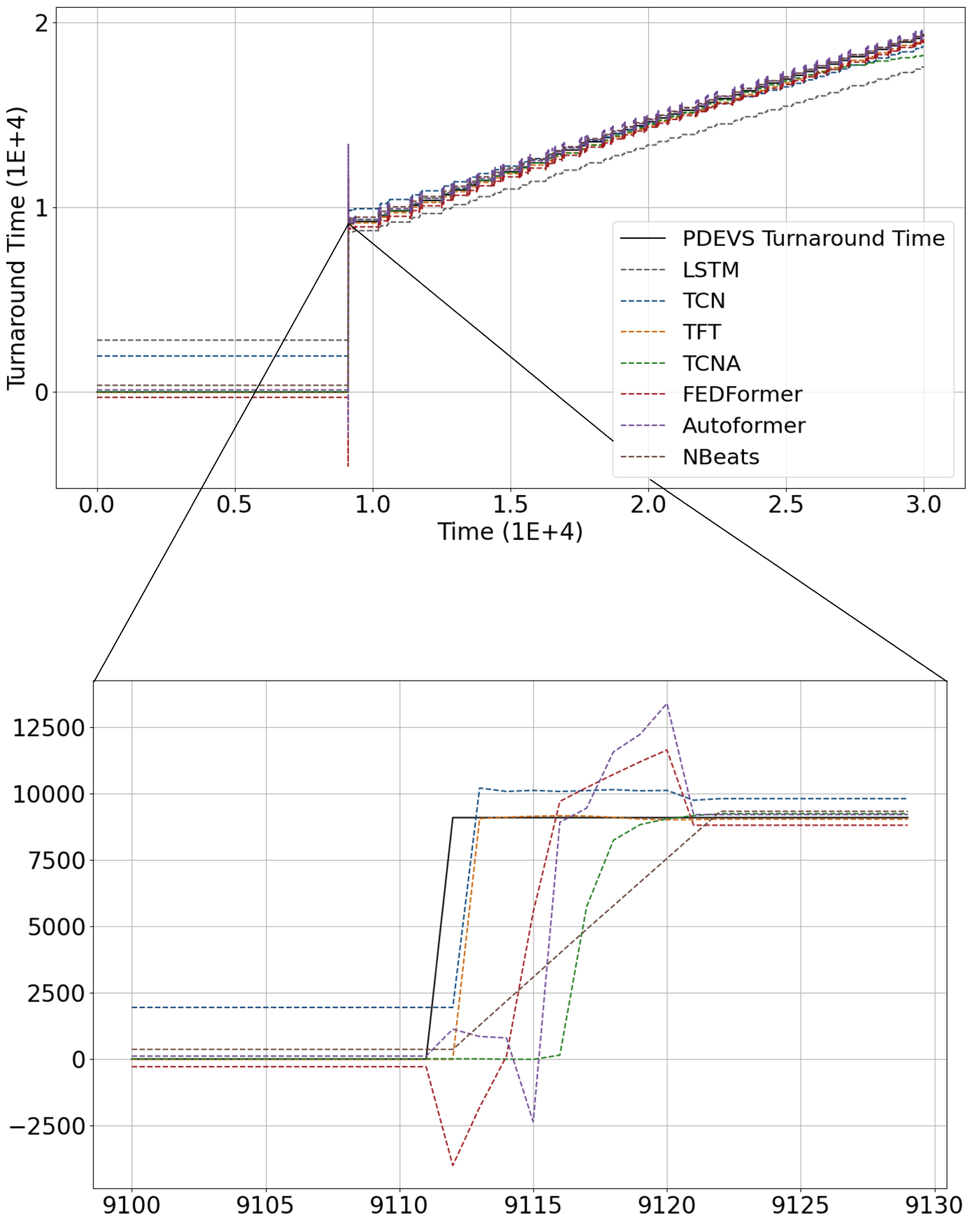}
    \caption{$TA$ Prediction Around the First Non-Zero Value for Medium Lot Size}
    \label{fig:turnaround_first_non_zero}
\end{figure}

Although throughput ($TH$) and turnaround time ($TA$) exhibit similar trends at a macroscopic level, their early- and late-stage prediction behavior reveals important differences among the temporal ML models. Figures~\ref{fig:throughput_first_non_zero} and~\ref{fig:turnaround_first_non_zero} illustrate model predictions around the first non-zero event, which corresponds to the time when the first wafer exits the MiniFab. This transition represents a sharp regime change that tests a model's ability to capture local temporal dynamics near a zero-valued boundary. While most models achieve strong aggregate performance in terms of $MAPE$, $MSE$, and $R^2$ (see Table~\ref{table:model_performance}), the early-stage behavior exposes inaccuracies that are not reflected in global metrics. As shown in Table~\ref{table:early-final-predictions},\footnote{Percentage differences are relative to the PDEVS reference values in Scenario E for the small (Pa = 3, Pb = 42, Tw = 15; lot size = 60), medium (Pa = 72, Pb = 45, Tw = 27; lot size = 144), and large (Pa = 150, Pb = 24, Tw = 18; lot size = 192) lot sizes. Early-horizon predictions correspond to model outputs after a 10-step look-back window (i.e., 10 minutes after the first non-zero event).} Several models exhibit large percentage deviations at the \emph{initial prediction step}, particularly for throughput. LSTM shows extremely high percentage differences (379\% to 1932\%), TFT, TCNA, and Autoformer deviate by approximately 87\% to 155\% for all lot sizes, and the transformer-based models without convolutional structure (Autoformer and FEDformer) exhibit significant variability. Notably, some models predict negative initial throughput values, which are physically infeasible in a manufacturing context and arise from unconstrained regression near zero-valued targets.

To further examine short-horizon stabilization effects, Table~\ref{table:early-final-predictions} also reports prediction errors after a \emph{look-back window of 10 time steps}, shown in the \textit{Early Horizon} rows. These rows quantify the percentage difference between the model predictions and the PDEVS values after the first 10 minutes of production. For most models, and particularly for FEDformer and Autoformer on turnaround time, this short historical context reduces the error relative to the initial step alone. However, large discrepancies remain for the throughput predictions of LSTM (875\% to 1907\%) and, to a lesser extent, TFT and TCNA (approximately 85\% to 99\%), whereas TCN attains the lowest early-horizon throughput errors (1.6\% to 13.6\%). For turnaround time, TCNA exhibits low percentage differences at both the initial step and the early horizon (0.49\% to 3.38\%) across all lot sizes, demonstrating rapid stabilization. The final-value comparisons further show that most models converge toward the PDEVS reference by the end of processing, which can mask early-stage inaccuracies if only terminal performance is considered. Overall, these results highlight the importance of jointly evaluating initial, short-horizon, and final prediction behavior, especially in semiconductor manufacturing applications, where early prediction accuracy is critical for operational decision-making.

\begin{table*}[!t]
\centering
\caption{Percentage Difference Between PDEVS and Model Predictions at Initial, Early-Horizon, and Final Time Steps for Throughput and Turnaround Time}
\label{table:early-final-predictions}
\resizebox{\textwidth}{!}{%
\begin{tabular}{cclllllllll}
\hline
\textbf{Timestamp} &
  \multicolumn{1}{l}{\textbf{}} &
  \textbf{Lot Size} &
  \textbf{PDEVS Values} &
  \multicolumn{7}{c}{\textbf{Percentage Difference from PDEVS Values}} \\ \hline
\multicolumn{1}{l}{\textbf{}} &
  \multicolumn{1}{l}{} &
   &
   &
  \textit{\textbf{LSTM}} &
  \textit{\textbf{TCN}} &
  \textit{\textbf{TFT}} &
  \textit{\textbf{TCNA}} &
  \textit{\textbf{FEDformer}} &
  \textit{\textbf{Autoformer}} &
  \textit{\textbf{N-BEATS}} \\ \hline
\multirow{6}{*}{\textbf{Initial}} &
  \multirow{3}{*}{\textit{\textbf{Throughput}}} &
  \textit{Small} &
  1.14E-04 &
  378.70\% &
  31.71\% &
  96.02\% &
  94.38\% &
  34.08\% &
  155.16\% &
  \textbf{3.21\%} \\
 &
   &
  \textit{Medium} &
  1.10E-04 &
  1931.99\% &
  28.55\% &
  87.50\% &
  97.23\% &
  32.03\% &
  100.16\% &
  \textbf{23.99\%} \\
 &
   &
  \textit{Large} &
  1.12E-04 &
  896.10\% &
  \textbf{12.20\%} &
  95.65\% &
  92.52\% &
  27.96\% &
  107.36\% &
  32.96\% \\ \cline{2-11} 
 &
  \multirow{3}{*}{\textit{\textbf{Turnaround Time}}} &
  \textit{Small} &
  8.75E+03 &
  13.28\% &
  \textbf{0.66\%} &
  3.47\% &
  1.85\% &
  28.10\% &
  49.26\% &
  20.54\% \\
 &
   &
  \textit{Medium} &
  9.09E+03 &
  8.06\% &
  11.29\% &
  0.92\% &
  \textbf{0.49\%} &
  28.05\% &
  47.30\% &
  16.87\% \\
 &
   &
  \textit{Large} &
  8.87E+03 &
  23.57\% &
  36.75\% &
  3.10\% &
  2.62\% &
  3.23\% &
  \textbf{1.29\%} &
  2.09\% \\ \hline
\multirow{6}{*}{\textbf{Early Horizon}} &
  \multirow{3}{*}{\textit{\textbf{Throughput}}} &
  \textit{Small} &
  1.14E-04 &
  315.60\% &
  \textbf{13.63\%} &
  97.48\% &
  90.70\% &
  62.52\% &
  24.59\% &
  51.14\% \\
 &
   &
  \textit{Medium} &
  1.10E-04 &
  1906.65\% &
  \textbf{10.21\%} &
  85.49\% &
  96.10\% &
  26.74\% &
  140.99\% &
  43.12\% \\
 &
   &
  \textit{Large} &
  1.12E-04 &
  874.67\% &
  \textbf{1.60\%} &
  99.34\% &
  91.60\% &
  63.40\% &
  24.94\% &
  52.07\% \\ \cline{2-11} 
 &
  \multirow{3}{*}{\textit{\textbf{Turnaround Time}}} &
  \textit{Small} &
  8.75E+03 &
  11.91\% &
  8.32\% &
  3.22\% &
  3.38\% &
  3.27\% &
  \textbf{1.30\%} &
  11.28\% \\
 &
   &
  \textit{Medium} &
  9.09E+03 &
  6.44\% &
  7.22\% &
  0.83\% &
  \textbf{0.81\%} &
  3.15\% &
  1.25\% &
  7.13\% \\
 &
   &
  \textit{Large} &
  8.87E+03 &
  22.39\% &
  38.94\% &
  2.52\% &
  1.64\% &
  3.23\% &
  \textbf{1.29\%} &
  11.34\% \\ \hline
\multirow{6}{*}{\textbf{Final}} &
  \multirow{3}{*}{\textit{\textbf{Throughput}}} &
  \textit{Small} &
  3.17E-03 &
  10.19\% &
  3.28\% &
  1.76\% &
  1.49\% &
  6.20\% &
  \textbf{0.45\%} &
  2.71\% \\
 &
   &
  \textit{Medium} &
  4.31E-03 &
  29.82\% &
  3.32\% &
  0.12\% &
  0.75\% &
  2.39\% &
  \textbf{0.09\%} &
  3.29\% \\
 &
   &
  \textit{Large} &
  4.60E-03 &
  3.53\% &
  0.50\% &
  0.84\% &
  1.48\% &
  3.18\% &
  \textbf{0.06\%} &
  2.54\% \\ \cline{2-11} 
 &
  \multirow{3}{*}{\textit{\textbf{Turnaround Time}}} &
  \textit{Small} &
  1.34E+04 &
  11.88\% &
  7.47\% &
  1.13\% &
  1.43\% &
  3.21\% &
  \textbf{1.04\%} &
  3.79\% \\
 &
   &
  \textit{Medium} &
  1.93E+04 &
  9.07\% &
  3.29\% &
  1.64\% &
  5.58\% &
  2.19\% &
  0.36\% &
  \textbf{0.14\%} \\
 &
   &
  \textit{Large} &
  1.89E+04 &
  4.94\% &
  12.55\% &
  0.59\% &
  2.45\% &
  2.19\% &
  \textbf{0.28\%} &
  4.29\% \\ \hline
\end{tabular}%
}
\end{table*}

\subsection{Efficiency Analysis}
\label{subsection:efficiency_analysis}

\begin{figure}[htb!]
    \centering
    \includegraphics[width=0.5\textwidth]{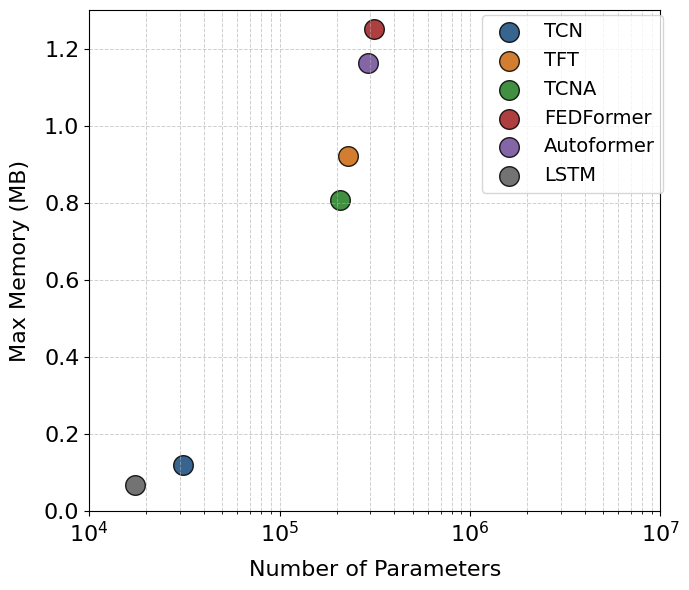} 
    \caption{Model Efficiency}
    \label{fig:efficiency_plot}
\end{figure}

The efficiency of the models is evaluated in terms of memory and the number of parameters needed for training. Like TFT, the TCNA model captures the essential information in the time series data. It needs less training time than all other models except LSTM and TCN, whose predictions are considerably less accurate (see Figure~\ref{fig:efficiency_plot}). The remaining temporal models also produce less accurate predictions. Compared with TFT, TCNA reduces the model size by approximately 10\%, the number of parameters by approximately 12\%, and the training time per epoch by approximately 9\%, which can be significant for longer sequence lengths $L$.

\subsection{Ablation Study Using Error Measurements}
\label{subsection:ablation_study}

We performed an ablation study to understand the importance of each part of the TCNA architecture and its impact on the overall accuracy of the model. Experiments were conducted by removing individual components of the MiniFab TCNA model, as shown in Table~\ref{table:ablation_study}. Removing the gating mechanism has the smallest adverse effect on accuracy, followed by removing the attention heads and then the initial TCN layer. Each component of the architecture therefore contributes to the overall accuracy and efficiency of the model. The gating mechanism contributes less to accuracy but is essential for maintaining the overall efficiency of the model.

\begin{table}[!htb]
\caption{Common Measurements for Ablation Study}
\label{table:ablation_study}
\resizebox{\columnwidth}{!}{%
\begin{tabular}{lccc}
\hline
\textbf{Method} & \multicolumn{1}{l}{\textbf{$MSE$}} & \multicolumn{1}{l}{\textbf{$R^2$ Score}} & \multicolumn{1}{l}{\textbf{$MAPE$}} \\ \hline
\textit{TCNA} & \textbf{2.60E-10} & \textbf{0.9995} & \textbf{0.86\%} \\
\textit{No Attention} & 5.02E-09 & 0.9984 & 3.21\% \\
\textit{No Initial TCN} & 2.60E-07 & 0.9170 & 21.94\% \\
\textit{No Gating Mechanism} & 1.56E-09 & \textbf{0.9995} & 1.17\% \\ \hline
\end{tabular}%
}
\end{table}

\section{Discussion}
\label{section:discussion}

The experimental results across all seven temporal ML models demonstrate that lot size and factory configuration play a critical role in shaping the temporal dynamics of both throughput ($TH$) and turnaround time ($TA$), directly influencing prediction accuracy and stability across models. Smaller lot sizes produce shorter processing horizons and sharper transitions, particularly at the point where the first wafers exit the 8-stage MiniFab; this amplifies local temporal dependencies and exposes the limitations of models that rely primarily on autoregressive behavior or long-horizon averaging. In such cases, LSTM and TCN often lag behind the ground-truth PDEVS simulation trajectories, while transformer-based models without convolutional structure tend to exhibit overshoot or oscillatory behavior. This early-stage sensitivity is further reflected in large percentage differences at the initial prediction step and, in some cases, physically infeasible negative throughput values, highlighting the challenge of modeling the PDEVS dynamics. These inaccuracies likely arise from the difficulty of approximating instantaneous state transitions in machines that concurrently receive and send wafers through multiple input and output ports. 

As the lot size increases, the throughput and turnaround time profiles become smoother and extend over longer horizons. Although this smoothing generally reduces short-term volatility and improves average error metrics for all models, it also introduces cumulative prediction bias, especially near saturation. This effect is evident in the large lot size configurations, where small temporal misalignments lead to noticeable drift in the $TA$ predictions. Incorporating short-horizon context via a look-back window reduces early-stage errors for most models; however, residual discrepancies remain for models with weaker short-term adaptability. TCNA mitigates many of these issues by combining convolutional feature extraction with attention-based contextual modeling, maintaining close alignment with the simulated PDEVS trajectories for turnaround time in both the early and late stages of processing, and for throughput once the initial transient has passed. Across small, medium, and large lot sizes, TCNA shows little variation in error behavior, indicating strong robustness to changes in batch configuration. TFT achieves competitive accuracy, and is more accurate than TCNA at several initial and final time steps, whereas TCNA attains comparable aggregate accuracy with a smaller model (Section~\ref{subsection:efficiency_analysis}). These findings suggest that explicitly modeling both local and global temporal dependencies is essential for better and reliable prediction in semiconductor fabrication manufacturing systems.

\section{Conclusion and Future Work}
\label{section:conclusion}

Semiconductor factories involve complex processes and feedforward/feedback flows for fabricating many kinds of chips. In these factories, wafers undergo many incremental processing steps spanning hours to weeks in connected machines with re-entrant flows. Essential to the operation of such factories is accurately predicting throughput ($TH$) and turnaround time ($TA$) dynamics for individual and multiple machines, repeatedly and as fast as possible (e.g., every hour). This paper presented a Temporal Convolutional Network with Multi-head Attention (TCNA) to predict these essential dynamics. Simulated data sets of a benchmark discrete-event semiconductor manufacturing factory, processing wafer lots with uniform and sinusoidal wafer arrival profiles and subject to preventive maintenance, were used to develop TCNA and the comparison models. The $TH$ and $TA$ time series predicted by TCNA, evaluated with $R^2$, $MSE$, and $MAPE$ and in terms of their dynamics at the beginning and end of wafer processing, show satisfactory accuracy with comparable parameter counts and memory utilization relative to the other temporal ML models.

Future work will extend this study in two directions. First, the evaluation will move beyond the point-accuracy measures used here ($MAPE$, $MSE$, and $R^2$) to a multi-metric assessment \cite{pendyala2026framework}: the mean absolute scaled error ($MASE$), which scales errors by the typical one-step change in the training data and remains well defined; accumulated and position-weighted errors (IAE, ISE, and ITAE), which reveal persisting and accumulated errors in long horizon trajectories; dynamic time warping (DTW) and slope errors, which assess the shapes of trajectories; peak and transition errors, which assess the timing of events; and frequency-domain errors. Such measures can rank models differently than point accuracy does and are better suited to the piecewise-constant state transitions, event-driven I/O trajectories of discrete-event simulation. Second, the architecture will be extended to dynamic covariates. The present model conditions only on static covariates (lot size, wafer-arrival profile, and repair type). The future-time branch shown dashed in Figure~\ref{fig:tcna} embeds known-future inputs, such as the event times of the steps to be predicted in a discrete-event trajectory, and joins them with the TCN summary and the static embedding in the output layer; time-varying inputs, such as the scaled simulation time, machine availability, or planned wafer releases, can likewise be added as further channels of the input window. Evaluating TCNA with these dynamic covariates and the multi-metric measures on discrete-event trajectories remains future work.

\end{document}